\pdfoutput=1
\documentclass{article}
\usepackage{iclr2027_conference,times}
\usepackage[T1]{fontenc}

\usepackage{amsmath,amsfonts,bm}

\def\eqref#1{equation~\ref{#1}}

\def\1{\bm{1}}

\DeclareMathAlphabet{\mathsfit}{\encodingdefault}{\sfdefault}{m}{sl}
\SetMathAlphabet{\mathsfit}{bold}{\encodingdefault}{\sfdefault}{bx}{n}

\usepackage{url}
\usepackage{graphicx}
\usepackage{booktabs}
\usepackage{array}
\usepackage{amsmath}
\usepackage{amssymb}
\usepackage{algorithm}
\usepackage{algpseudocode}
\usepackage{hyperref}

\graphicspath{{figures/}}

\title{Vestrum: Learning the Verification, \\ Retrieval and Structure of Agent Harnesses \\ from a Few Long Episodes}

\author{Jayant Parashar$^{1}$, Eugene F. Douglass$^{1}$, William C. Bastian$^{2}$,
Suchendra M. Bhandarkar$^{1}$ \\
\normalfont $^{1}$University of Georgia \qquad $^{2}$Emory University
}
\hypersetup{
  pdftitle={Vestrum: Learning the Verification, Retrieval and Structure of Agent Harnesses from a Few Long Episodes},
  pdfauthor={Jayant Parashar, Eugene F. Douglass, William C. Bastian, Suchendra M. Bhandarkar}
}

\iclrfinalcopy

\begin{document}

\maketitle
\lhead{Preprint. Under review.}

\begin{abstract}
Agent harnesses decide what the model retrieves, decomposes, remembers and verifies, and most
people use harnesses that are pre-built and fixed, such as Claude Code or a published memory
system. Harness optimizers rewrite their own harness code rather than editing pre-built systems,
and they learn from tens to hundreds of tasks, while long-horizon work offers a handful of
episodes, each hundreds of tool calls long. We present Vestrum, which improves
existing harnesses component by component. It reads full raw traces, states each failure as a
recognizable class, and changes the one component where the error arose (a self-verification, a
retrieval view, a decomposition step or a memory operation). Each change targets a failure that recurs across tasks of
the same kind. Before anything runs, a generalization filter refuses changes that copy facts
from the development tasks. Vestrum learns from one to ten long development episodes and improves
tasks it has never seen in all five settings. On Claude Code, Vestrum raises Terminal-Bench~4 Hard
from 63.7\% to 70.3\% of test checks at $1.03\times$ the test cost. It raises expert cell-type agreement
from 67.5\% to 77.8\% on held-out sections of one slide. On the GAM memory system, it raises
UltraHorizon from 47.6 to 59.8, AMA-Bench from 64.2\% to 74.2\% and LoCoMo from 71.5\% to 74.9\%.
It beats the GEPA prompt search we ran on the same data in every paired evaluation, and these gains
hold under a test model different from the one Vestrum's search used. One loop learns self-verification,
retrieval and structure. Across the searches, every self-verification the loop kept acts early: it
checks an intermediate result against recorded evidence before the result is committed, or, on
Terminal-Bench, requires an invariant and a probe before the first edit. In development, auditors that re-examined every finished answer broke more correct
answers than they fixed, and the loop did not keep them.
\end{abstract}

\section{Introduction}
\label{sec:introduction}

An agent harness is the software around a language model. It decides what the model verifies
before a result is used, what it retrieves and in what form, how a task is divided, and what is
recorded for later. Many harnesses in use were developed by someone other than the people who deploy
them. Claude Code is a closed coding agent. General Agentic Memory (GAM) is a published
memory system: its memorizer records an incoming stream, and its researcher
retrieves from the resulting store \citep{gam2025}. Improving such a harness means changing
some of its components while the others still work.

Harness optimizers instead rewrite harness code they own. Meta-Harness gives a proposer the code, scores and raw
execution traces of every earlier candidate and searches over whole harness programs
\citep{lee2026metaharness}. Agentic Harness Engineering (AHE) grows a harness from a minimal seed
\citep{lin2026ahe}. ADAS, AFlow, the Darwin G\"odel Machine and self-improving coding agents
generate or rewrite agents they own
\citep{hu2024adas,zhang2025aflow,zhang2025dgm,robeyns2025sica}. They learn from tens to hundreds
of tasks: Meta-Harness and AHE from the 89 Terminal-Bench~2 tasks, on which they also report
their results. Long-horizon work offers far fewer. Each episode runs for hundreds of tool
calls, and early mistakes propagate \citep{luo2025ultrahorizon}. A benchmark provides a handful
of such episodes. With so few development tasks, the cheapest way to raise their score is to
copy their answers, names or paths into the harness.

Vestrum improves existing harnesses component by component, from one to ten long episodes. It
uses two earlier ideas: raw traces contain the information a harness proposer needs
\citep{lee2026metaharness}, and agent failures fall into recognizable classes
\citep{cemri2025mast}. A language model examines each failed trace, finds where the error began and
states it as a \emph{failure class}. The proposer then changes the one component at that point;
this change is a \emph{mechanism}. Each mechanism belongs to one of four \emph{design dimensions}:
self-verification, retrieval, decomposition and knowledge synthesis. Decomposition divides the work
and knowledge synthesis decides what is distilled and recorded; these two set the harness's structure.
Each mechanism targets a failure that recurs across tasks of the same \emph{task kind}. The rest of the
\emph{base harness} that Vestrum starts from runs unchanged. The \emph{generalization
filter} is a language-model reviewer and a lexical lint for task-specific strings.
Before anything runs, it refuses any mechanism that states a fact about the development tasks
instead of addressing its failure class. The loop then runs the remaining mechanisms together as
one \emph{bundle} on the development tasks. It admits the bundle only if the paired comparison
meets the admission criteria. One pass of this loop is an \emph{epoch}.

The learned harnesses, fixed before testing, improve held-out tasks in all five settings, on both
base harnesses. On Claude Code, Vestrum raises Terminal-Bench~4 Hard by 6.6 points of test checks
at 1.03 times the test cost. It raises expert cell-type agreement on PathBench by 10.3 points. On
GAM, it raises UltraHorizon by 12.2 points, AMA-Bench by 10.0 and LoCoMo by 3.4. It beats a GEPA
prompt search run on the same development data in every paired evaluation, and the memory gains
transfer to a test model that Vestrum's search never used. It does this cheaply: one to ten
development episodes per setting, \$123.5 for the reported Terminal-Bench epoch including its test
runs, and an estimated quarter of GEPA's task executions on LoCoMo and AMA-Bench. AMA-Bench and
PathBench used no validation split and still improved by 10.0 and 10.3 points. With so few
episodes, a mechanism that copies development content is easy to produce, and the filter refused
such mechanisms. On PathBench its reviewer refused 5 of 19 proposals as specific to the
development sections.

One loop learns self-verification, retrieval and structure together. Across the searches it kept
some self-verifications and not others, and the ones it kept share a placement. On the memory
benchmarks, every self-verification it kept compares an intermediate result with recorded evidence
where that result is committed: a comparison with the source chunk at the memory write on
AMA-Bench, a claim judge at the evidence merge on LoCoMo, a commit gate at the rule commitment on
UltraHorizon. On Terminal-Bench, an edit gate requires an invariant and a probe before the first
edit of a shipped file. In development, auditors that re-examined every finished answer broke more
correct answers than they fixed: 9 against 5 on AMA-Bench and 6 against 2 on LoCoMo. The loop did
not keep them.

\paragraph{Contributions.}
\begin{itemize}
  \item \textbf{Component-wise harness learning.} In a harness it does not own, Vestrum changes the
  one component where an error arose, for a failure that recurs within a task kind. It
  improves a closed coding agent (Claude Code) and a published memory system (GAM) without
  rewriting either, at 1.03 times Claude Code's test cost (Sections~\ref{sec:method}
  and~\ref{sec:experiments}).
  \item \textbf{A generalization filter for learning from a few episodes.} Before anything runs, a
  reviewer and a lexical lint refuse mechanisms that encode the development tasks. On
  PathBench the reviewer refused 5 of 19 proposals. Vestrum learns from one to ten long episodes
  and improves held-out tasks in all five settings, two of them without a validation split
  (Sections~\ref{sec:population} and~\ref{sec:experiments}).
  \item \textbf{One loop over self-verification, retrieval and structure.} The learned harnesses
  change all four design dimensions. The search admitted evidence-grounded self-verification at the
  memory write, the evidence merge and the rule commitment, and an edit gate at the first edit. On
  measured fix and break counts, the loop did not keep auditors that re-examined every finished
  answer (Section~\ref{sec:architectures}, Appendix~\ref{app:verification}).
\end{itemize}

\section{Related Work}
\label{sec:related}

\paragraph{Harness and agent search.}
Automated design of agentic systems searches program space \citep{hu2024adas,zhang2025aflow},
and self-improving agents rewrite their own source
\citep{zhang2025dgm,robeyns2025sica,yin2024godel}. Meta-Harness searches whole harness programs
from raw execution traces \citep{lee2026metaharness}, and AHE links
localized failures to file-level edits of a harness it grows from a minimal seed
\citep{lin2026ahe}. Each of these edits code it owns, learns from tens to hundreds of tasks, and
guards generalization with validation scores or, in Meta-Harness, a regular-expression audit
after search. Life-Harness adapts the interface of deterministic tool agents through its own
wrapper \citep{xu2026lifeharness}. SkillOpt runs inside Claude Code, edits a single skill
document and accepts an edit only if a validation score improves \citep{yang2026skillopt}.
Vestrum differs in all three respects: it changes individual components of harnesses it does not
own, it learns from one to ten long episodes, and it filters each mechanism
for task content before any execution.

\paragraph{Prompt, program and memory optimization.}
DSPy, MIPRO, TextGrad, ACE and GEPA optimize prompts and programs
\citep{khattab2023dspy,opsahlong2024mipro,yuksekgonul2024textgrad,zhang2025ace,agrawal2025gepa}.
GEPA reflects on trajectories and maintains an instance-level Pareto front. A prompt cannot add
a memory operation. On LoCoMo, M$^\star$ finds prompt-only evolution far behind program evolution
\citep{pan2026mstar}. ALMA searches memory mechanism code, and EvolveMem tunes retrieval settings
\citep{alma2026,evolvemem2026}. Vestrum changes the memory operations, views and control flow of
an existing memory system, and it beats GEPA on three memory benchmarks. Memory systems themselves
page context against external storage \citep{memgpt2023}, construct notes, graphs and indexes
\citep{amem2025,zep2025,hipporag2024,chhikara2025mem0}, or organize information per query
\citep{gam2025}.

\paragraph{Failures and self-verification.}
Failure taxonomies and attribution \citep{cemri2025mast,zhang2025whowhen,zhu2025agentfail} give
Vestrum its failure classes. Reflexion, ExpeL, AutoGuide and Agent Workflow Memory give it
the idea of lessons kept across attempts
\citep{shinn2023reflexion,zhao2024expel,fu2024autoguide,wang2024awm}. Process
supervision grades intermediate steps \citep{uesato2022process,lightman2024verify}, and Sherlock
attaches verifiers to error-prone nodes of a declared workflow \citep{ro2025sherlock}.
Self-correction without external feedback turns correct answers into wrong ones
\citep{huang2024selfcorrect,kamoi2024selfcorrect}. From its own traces, Vestrum learns where to
place self-verification mechanisms. The final-answer auditors it did not keep reproduce the
self-correction result inside a harness.

\paragraph{Concurrent work.}
Several harness learners posted while Vestrum was being completed arrive independently at
related ideas. HarnessFix maps recurring flaws to scoped repair operators
\citep{chen2026harnessfix}. AdaMAST passes taxonomy-coded diagnoses to its mutator and also
improves Claude Code \citep{cemri2026adamast}. ModularRSI restricts each edit to one module
\citep{wu2026modularrsi}. HarnessEvolve, HarnessCompass and RRSI add language-model leakage
gates or critics, each alongside validation or regression sets and 50 or more tasks
\citep{jiang2026harnessevolve,zhang2026harnesscompass,xia2026rrsi}. RHO optimizes an add-on folder
from a ten-task coreset \citep{pan2026rho}. Self-Harness, Harness-R1 and AutoSaddler revise
harnesses from failure traces \citep{zhang2026selfharness,shao2026harnessr1,park2026autosaddler},
and MemPro evolves memory systems as programs \citep{mempro2026}. Two studies show why the
few-task regime needs a guard: evolved edits often memorize fixes \citep{wang2026rethinking}, and
whole-codebase evolution from five ARC-AGI-3 games yields small held-out gains
\citep{lehmann2026arcagi3}. Table~\ref{tab:neighbours} in Appendix~\ref{app:related} sets Vestrum
beside the closest of these.

\section{Learning Filtered Harness Mechanisms}
\label{sec:method}

\begin{figure}[t]
\centering
\includegraphics[width=0.9\linewidth]{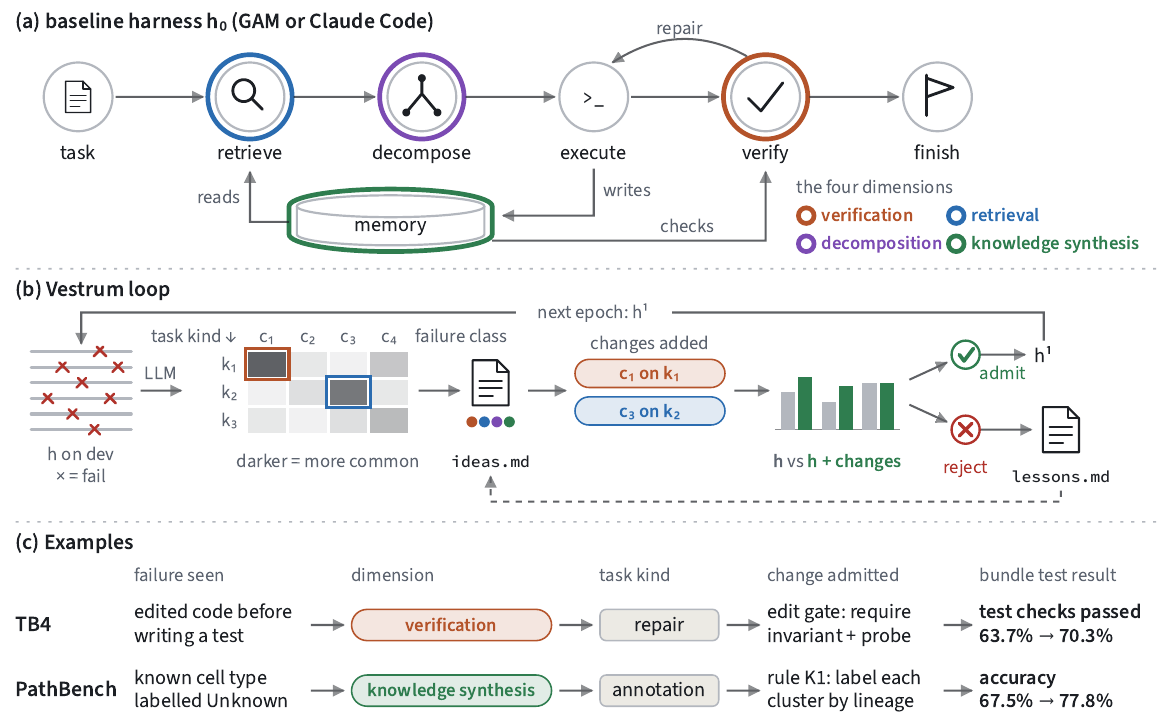}
\caption{\textbf{(a)} The four design dimensions within an agent harness.
\textbf{(b)} Development failures are organized by task kind and failure class; darker cells
indicate recurrence across more tasks. Scoped mechanisms are composed, filtered and
evaluated, and lessons pass to later epochs.
\textbf{(c)} Two example mechanisms and the results of their containing harnesses.}
\label{fig:vestrum}
\end{figure}

Vestrum maintains a current harness and a lessons file. In each epoch it examines development
traces, states their failures as classes and places a mechanism for each class at one component.
It filters the mechanisms and runs those that pass together as one bundle. The harness changes only
when the bundle meets the admission criteria (Figure~\ref{fig:vestrum};
Algorithm~\ref{alg:vestrum}). All five settings run the same loop. Each benchmark provides an
interface for running, scoring and categorizing tasks.

\subsection{Search state, task kinds and failure classes}
\label{sec:setting}

The search starts from the base harness $h_0$ (GAM or Claude Code). It maintains the current
harness $h$ and the lessons file $L$. Within a search the task model, grader and benchmark checks remain
fixed, and no model is trained. Development tasks provide traces and outcomes. Where validation
is used, it is disjoint from development and test. Test outcomes never return to the loop.

Task kinds are assigned before any proposal, from task descriptions or benchmark categories. On
Terminal-Bench a task kind combines the task mode, such as repair or build, with the check style
described in the instruction. On the memory benchmarks a task kind is an environment, question
category or probe axis. A task kind states where a mechanism should transfer, and a failure class states
what recurs there. Neither is redefined to justify a proposal.

\subsection{From traces to recognizable failures}
\label{sec:observations}

Each failed check marks a development outcome to explain. A check is a benchmark test or a
judged probe. Building on structured failure analysis \citep{cemri2025mast}, a language model
examines the trace, localizes the apparent failure and expresses it as a predicate recognizable
in other traces. For example, a repair agent may change a shipped file before
establishing the invariant the code must preserve. Here \emph{repair} is the task kind and
\emph{changing code before establishing the invariant} is the failure class. The predicate
describes behavior and prescribes no mechanism, so several mechanisms may address one class.

\subsection{Placing a mechanism at a component}
\label{sec:change}

A mechanism can go in one of the four design dimensions: self-verification (what
the harness verifies before a result is used), retrieval (what the agent is shown and in what form), decomposition (how
work is divided) and knowledge synthesis (what the harness distills and records). For each
failure class, the proposer is given the lessons and the localized failure and places the
mechanism where the failure arises. In the repair example that place is the first edit. The
mechanism is an \emph{edit gate}, which requires an invariant and a probe script before the first
change to a shipped file. A mechanism may change a prompt, a tool, a memory operation or the
control flow.

Mechanisms attach to the base harness without rewriting it. On Claude Code, Vestrum delivers
them as instruction files and prompts before the first turn and enforces gates around the
agent's actions, so it needs no access to Claude Code's source. On GAM, a mechanism adds or
replaces one memory operation, view or prompt. Everything else in the base harness runs
unchanged.

Each proposal states its target failure class, applicable task kinds, recognition predicate and
predicted check improvements. A language model scores the proposals and gives priority to classes
seen on more tasks. The most promising mechanisms are composed into a bundle $S$. $h+S$ denotes
the current harness with that bundle applied. Appendix~\ref{app:loop} describes the combined composer used on Terminal-Bench.

\subsection{The generalization filter and scope}
\label{sec:population}

With few development tasks, a mechanism can raise their score by encoding them. Before any
execution, the generalization filter tests whether a mechanism addresses its failure class or
states a fact about the development tasks. A language-model reviewer examines the mechanism's
condition and content, and a lexical lint rejects task-specific names, symbols, paths and
constants. Only mechanisms that pass the filter are run, so a refused mechanism costs no episode.
On Terminal-Bench, for example, the edit gate fires on repair tasks and on no other request. Appendix~\ref{app:supervision} lists the supervision each
setting permits.

\subsection{Bundle evaluation and admission}
\label{sec:admission}

On development tasks, the loop evaluates $h+S$ against the current harness, reusing existing
runs of $h$ where available. Trace analysis then marks mechanisms that did not act or that
looked neutral or harmful. A pruned bundle $S'$ drops or repairs them. If $S'$ differs from
$S$, $h+S'$ is also evaluated. The diagnoses guide refinement, but admission depends only on
the measured performance of an evaluated bundle. Where validation is used, development results
decide which candidates advance to it. A candidate replaces $h$ only if it meets the admission
criteria, including validation where used; otherwise $h$ remains. The criteria are set per
setting before its search (Table~\ref{tab:implementation}). A \emph{draw} is one run of a task. On
Terminal-Bench, for example, a bundle must pass checks that Claude Code fails in every
draw, lose no check on a task of another kind, and cost under
$1.5\times$ the baseline per task.

The \emph{ceiling probe} is a Terminal-Bench run that gives the agent each development task's solved
procedure; proposals then target only the tasks where this run passes checks Claude Code fails. In one ablation search without it, scope decided admission. A mechanism
that added reference notes on every task was not admitted after a task of another kind lost a
check, and the same mechanism restricted to repair tasks was admitted.

\subsection{Learning across epochs}
\label{sec:loop}

The lessons file records successes, failures and repeated refusals for later proposals, and it
is separate from task memory. It is written at the end of each epoch. Each epoch starts from the retained harness and the updated
lessons file (Algorithm~\ref{alg:vestrum}). The final harness is fixed before test evaluation.

\begin{algorithm}[t]
\caption{Vestrum: one epoch of harness adaptation. $R$ holds execution traces and scores;
\textsc{Filter} is the generalization filter; \textsc{AdmitBest} applies the admission
criteria and optional validation, and returns $h$ if no candidate qualifies.}
\label{alg:vestrum}
\small
\begin{algorithmic}[1]
\Require harness $h$; development tasks $D$; task kinds $K$; lessons file $L$
\Require optional validation tasks $V$
\State $R_0 \gets \Call{EvaluateOrReuse}{h,D}$
\State $F \gets \Call{Diagnose}{R_0,K}$
\State $\mathcal{D} \gets \{\text{self-verification, retrieval, decomposition, knowledge synthesis}\}$
\State $P \gets \Call{Propose}{F,K,L,\mathcal{D}}$
\State $S \gets \Call{RankAndCompose}{P};\quad S \gets \Call{Filter}{S}$
\State $R \gets \Call{Evaluate}{h+S,D}$
\State $\mathcal{C} \gets \{(h+S,R)\}$
\State $S' \gets \Call{Prune}{S,R_0,R}$
\If{$S' \ne S$}
  \State $R' \gets \Call{Evaluate}{h+S',D}$
  \State $\mathcal{C} \gets \mathcal{C} \cup \{(h+S',R')\}$
\EndIf
\State $h' \gets \Call{AdmitBest}{h,R_0,\mathcal{C},V}$
\State $L' \gets \Call{UpdateLessons}{L,F,\mathcal{C},h'}$
\State \Return $(h',L')$
\end{algorithmic}
\end{algorithm}

\section{Experiments}
\label{sec:experiments}

We run Vestrum in five settings (Table~\ref{tab:settings}). Terminal-Bench~4 Hard and
PathBench start from Claude Code. UltraHorizon, LoCoMo and AMA-Bench start from GAM, with A-Mem
as a further comparator \citep{gam2025,amem2025}. Development splits range from one conversation
to ten episodes. Table~\ref{tab:implementation} gives each setting's editable surface, task
kinds and admission criteria. UltraHorizon, LoCoMo and Terminal-Bench use a validation split
disjoint from development and test. AMA-Bench and PathBench use none, so their harnesses were
fixed on development results alone. Each setting has one search and one harness
fixed before testing, and test repetitions re-run it.

\subsection{Terminal-Bench~4 Hard}
\label{sec:tb4}

Terminal-Bench grades software tasks with tests hidden from the agent
\citep{merrill2026terminalbench}. During both search and test,
Claude Sonnet~5 executes the tasks. Neither harness passes every check on any held-out task,
so we report the share of checks passed. On eight held-out tasks with two draws per harness,
Vestrum raised checks passed from 63.7\% to 70.3\%. Its test cost was $1.03\times$
Claude Code's under the same per-task spend cap. Four task means improved, four were
unchanged, and none declined (Table~\ref{tab:tb4-test8}). The learned harness classifies each
request by task kind and provides four scoped mechanisms before the first turn
(Table~\ref{tab:components}).

\begin{table}[t]
\caption{Terminal-Bench~4 Hard: test checks passed on the eight test tasks, mean over two
draws per harness.}
\label{tab:tb4-test8}
\centering
\small
\begin{tabular}{lcccc}
\toprule
task & checks & Claude Code & Vestrum & $\Delta$ \\
\midrule
hof-topology-interpenetration & 38 & 21.5 & 25.0 & $+3.5$ \\
bun-sourcemap-leak           & 36 & 24.0 & 26.5 & $+2.5$ \\
cargo-flight-dispatch        & 27 & 17.0 & 19.0 & $+2.0$ \\
sglang-qwen-burst            & 13 & 5.0  & 7.0  & $+2.0$ \\
vllm-deepseek-streaming      & 5  & 1.0  & 1.0  & 0 \\
vpp-loss-divergence          & 5  & 4.0  & 4.0  & 0 \\
glycan-ms2-elucidation       & 12 & 11.0 & 11.0 & 0 \\
photonic-waveguide-routing   & 14 & 12.0 & 12.0 & 0 \\
\midrule
total                        & 150 & 95.5 & 105.5 & $+10.0$ \\
\bottomrule
\end{tabular}
\end{table}

\subsection{Agentic memory: UltraHorizon, LoCoMo and AMA-Bench}
\label{sec:memory}

The three memory benchmarks use retained information in different ways. In UltraHorizon the
agent must discover environment rules over long interactions \citep{luo2025ultrahorizon}.
LoCoMo poses questions about long conversations \citep{locomo2024}, and AMA-Bench poses
questions about tool-using trajectories \citep{zhao2026amabench}. Claude Opus~5 ran the search. The task model
during the search was Claude Haiku~4.5 on UltraHorizon and LoCoMo and GPT-4o-mini on
AMA-Bench. For testing, all memory systems used Gemini~3.7 Flash. We report \emph{judge accuracy} on
LoCoMo and AMA-Bench: the share of answers that a blinded judge grades fully
correct against the reference answer's key facts, by majority over three passes
(Appendix~\ref{app:judge}). The adversarial, unanswerable category of LoCoMo is excluded
(Appendix~\ref{app:memory-setup}).

\begin{table}[t]
\caption{Memory benchmark performance with Gemini~3.7 Flash ($\pm$: standard deviation).
Upper block: test repetitions of each final harness; the AMA-Bench row is the second
answer-generation run, and the difference row gives both runs. Lower block: a separate
comparison in which GAM, GEPA, and Vestrum were rejudged together, so GAM and Vestrum scores
differ slightly from the upper block; GEPA differences are computed within it.}
\label{tab:memory}
\centering
\small
\begin{tabular}{lccc}
\toprule
 & UltraHorizon & LoCoMo & AMA-Bench \\
 & score / 100 & judge acc. & judge acc. \\
\midrule
A-Mem   & 45.7 & $49.3 \pm 0.6\%$ & 56.2\% \\
GAM     & 47.6 & $71.5 \pm 0.6\%$ & 64.2\% \\
\textbf{Vestrum} & \textbf{59.8} & $\mathbf{74.9 \pm 0.8\%}$ & \textbf{74.2\%} \\
Vestrum $-$ GAM & $+12.2$ (3/3 seeds) & $+3.4$ (3/3 replicates) & $+12.1$, $+10.0$ (2 runs) \\
\midrule
\multicolumn{4}{l}{\emph{GEPA comparison}} \\
GAM & $47.6 \pm 0.7$ & $72.5 \pm 0.7\%$ & $65.7 \pm 0.9\%$ \\
GEPA on GAM & $47.9 \pm 2.0$ & $74.0 \pm 0.8\%$ & $70.8 \pm 1.5\%$ \\
\textbf{Vestrum} & $\mathbf{59.8 \pm 4.7}$ & $\mathbf{76.3 \pm 0.7\%}$ & $\mathbf{76.4 \pm 0.9\%}$ \\
Vestrum $-$ GEPA & $+11.9$ (3/3 seeds) & $+2.3$ (3/3 seeds) & $+5.6$ (3/3 judging runs) \\
\bottomrule
\end{tabular}
\end{table}

\paragraph{Gains over GAM.}
On all three benchmarks the learned harness improves on GAM (Table~\ref{tab:memory}).
UltraHorizon rose from 47.6 to 59.8, and the difference was positive at every seed. LoCoMo
rose from 71.5\% to 74.9\%. Across three replicates, the net gains were $+24$, $+39$ and
$+34$ probes. In two generation runs, AMA-Bench improved by $+12.1$ and $+10.0$ points, with
the largest margin on state abstraction. The paired records are in
Appendices~\ref{app:uh-seeds}--\ref{app:ama-seeds}.

\paragraph{Comparison with GEPA.}
GEPA evolves prompts by reflecting on traces \citep{agrawal2025gepa}. We ran it on the same
development data, with Gemini~3.7 Flash (the test model) as the task model, Claude Opus~5 for
reflection, and GAM's question-time prompts as its editable surface. Vestrum beat GEPA in every
paired evaluation: by 11.9 points on UltraHorizon, 5.6 on AMA-Bench and 2.3 on LoCoMo. GEPA's
search over 60 episodes on UltraHorizon found no prompt better than GAM's original. Vestrum also changes memory operations, tools and control flow, and on
AMA-Bench its margin over GEPA comes with deterministic trajectory views, which a prompt
cannot add. Vestrum's estimated task executions were 0.28 and 0.25 times GEPA's on LoCoMo and
AMA-Bench and equal on UltraHorizon, although it used more learning tokens on LoCoMo and
UltraHorizon (Appendix~\ref{app:gepa}).

\subsection{PathBench: reconstructing expert cell-type annotations}
\label{sec:pathbench}

PathBench requires an agent to reconstruct expert cell-type annotations from spatial
transcriptomics measurements. The data are six tissue sections of one head-and-neck squamous
cell carcinoma CosMx slide from the AnchorR study \citep{morris2026anchorr}. The
reference labels come from a bioinformatician's Seurat-based pipeline. Given the measurement
export, seven published reference sources and two in-house reference files, the agent must recover the sections, filter and
cluster cells, and assign labels. Two sections are used for development and the other four
for test. No test label or measurement was read before the harness was fixed.
Claude Opus~5 executes tasks. Because PathBench has a single task kind, it tests the
knowledge-synthesis dimension of the loop and does not test scope.

In development traces, the agent left clusters unlabelled despite evidence for their lineage.
From the labelled development sections, Vestrum learned eleven sourced annotation rules, marker
and expression reference tables, and a nearest-neighbour helper. All of these are provided before
the agent plans. On the held-out sections, agreement with the
expert labels rose from 67.5\% to 77.8\% and macro-F1 from 64.8 to 76.5, over two runs per
harness (Table~\ref{tab:pathbench}). Cell matching and scoring are described in
Appendix~\ref{app:pathbench}.

\begin{table}[t]
\caption{PathBench performance against expert annotations, averaged across execution runs ($n$).}
\label{tab:pathbench}
\centering
\small
\begin{tabular}{llccccc}
\toprule
split & system & $n$ & ARI & NMI & accuracy & macro-F1 \\
\midrule
development & Claude Code & 3 & 0.582 & 0.662 & 66.5\% & 59.0\% \\
            & Vestrum & 2 & 0.670 & 0.717 & 82.2\% & 80.4\% \\
\midrule
test & Claude Code & 2 & 0.642 & 0.713 & 67.5\% & 64.8\% \\
             & \textbf{Vestrum} & 2 & \textbf{0.684} & \textbf{0.725} & \textbf{77.8\%} & \textbf{76.5\%} \\
\bottomrule
\end{tabular}
\end{table}

\subsection{Ablating the loop}
\label{sec:ablation}

\begin{table}[t]
\caption{Terminal-Bench~4 Hard: ablation of the loop. The second row runs the third epoch without the lessons from the first
two. Below the line, parts of the loop are removed cumulatively: each row
also lacks those removed above it. A gain is a development check Claude Code fails in every draw and the candidate passes in every
draw; a loss is the reverse. Cost is task spend relative to Claude Code.}
\label{tab:ablation}
\centering
\footnotesize
\setlength{\tabcolsep}{4pt}
\begin{tabular}{@{}>{\raggedright\arraybackslash}p{3.1cm}>{\raggedright\arraybackslash}p{5.2cm}ccc@{}}
\toprule
loop & mechanisms found & gain & loss & cost \\
\midrule
full loop & repair procedures; edit gate; scale decision & $+16$ & 1 & $1.32\times$ \\
$-$ lessons file & all other loop parts retained & $+10$ & 0 & $1.6\times$ \\
\midrule
$-$ ceiling probe & scoped reference notes; fixture rule & $+3$ & 0 & $0.86\times$ \\
$-$ failure classes, generalization filter & adversarial checks; non-author reviewer & $+2$ & 0 & $1.7\times$ \\
$-$ localization & constraint checks; pre-planning probe & $+9$ pts$^{\dagger}$ & -- & $2.3\times$ \\
$-$ trace access (final outputs only) & audit checklist; two-phase audit & $+1$ & 0 & $2.1\times$ \\
\bottomrule
\end{tabular}
\par\vspace{2pt}
{\scriptsize $^{\dagger}$Percentage points of mean check share from single draws; individual checks cannot be counted.}
\end{table}

Table~\ref{tab:ablation} removes parts of the loop on Terminal-Bench, with one search per
variant. Appendix~\ref{app:earlier-loops} gives each variant's run. The full loop gained 16 development
checks and lost one. Without the lessons from the first two epochs it gained 10. The ceiling probe picks out failures a procedure can fix; without
it, the loop gained 3. Removing the generalization filter, failure classes and scope as well left 2 checks at
1.7 times the baseline's cost. That variant's first mechanism cost two checks on a task of another
kind, and three checks it gained on one development task did not reproduce in the next draw.

\section{What the Loop Learned}
\label{sec:architectures}

Table~\ref{tab:components} lists the admitted mechanisms by design dimension. Each setting's
bundle was learned separately; Appendix~\ref{app:components} describes each mechanism. The UltraHorizon and LoCoMo bundles change all four design dimensions. Most mechanisms change what no prompt can reach:
notebooks built at ingestion, event and object indexes, a commit gate before a rule is committed, and an
edit gate enforced around the agent. The Terminal-Bench repair procedures and the LoCoMo answer
rules are the textual exceptions.

\begin{table}[t]
\caption{Admitted mechanisms by design dimension. A dash means no admitted mechanism in that
dimension.}
\label{tab:components}
\centering
\small
\setlength{\tabcolsep}{4pt}
\begin{tabular}{@{}>{\raggedright\arraybackslash}p{0.15\linewidth}>{\raggedright\arraybackslash}p{0.18\linewidth}>{\raggedright\arraybackslash}p{0.17\linewidth}>{\raggedright\arraybackslash}p{0.19\linewidth}>{\raggedright\arraybackslash}p{0.22\linewidth}@{}}
\toprule
Setting & Self-verification & Retrieval & Decomposition & Knowledge synthesis \\
\midrule
UltraHorizon & Replay evidence before commitment & Pin notebooks by rule class & Re-analyze rules; plan probes & Record evidence with provenance \\
LoCoMo & Verify claims at evidence merges & Retrieve again when evidence is insufficient & Split questions selectively & Prioritize evidence; give a best-supported answer \\
AMA-Bench & Verify notes against source records & Index events and object histories & -- & Build notebooks during ingestion \\
PathBench & -- & -- & -- & Provide annotation rules and reference tables; reassign small unlabelled clusters \\
Terminal-Bench & Require invariant and probe before edits & -- & -- & Provide repair procedures; choose raw or log scale \\
\bottomrule
\end{tabular}
\end{table}

\subsection{Where self-verification belongs}
\label{sec:verification}

Self-verification is one of the four design dimensions, so the loop could place it at any point of a
run, or nowhere. Table~\ref{tab:verification-full} lists the admitted self-verification
and the auditors the loop did not keep. On the memory benchmarks, each admitted
self-verification compares a result with
its evidence. On
AMA-Bench each note is verified against its source chunk when it is written. In
development it repaired 17 of the 18 notes it examined. On LoCoMo a claim judge at the
evidence merge scored best of three placements (0.796 against 0.763--0.770; the placements also
differ in routing, Appendix~\ref{app:verification}). The UltraHorizon
commit gate replays logged observations before a rule is committed, and the Terminal-Bench edit gate requires
an invariant and a probe before the first edit. Switching these checks off at test lowered the score at
every seed on UltraHorizon grid episodes (70.8 to 64.1; GAM 56.5) and on LoCoMo (75.1\% to 73.2\%;
GAM 71.5\%; Appendix~\ref{app:verification-ablation}).

Auditors that re-examined every finished answer broke more correct
answers than they fixed. On AMA-Bench one auditor rewrote 82 of 120 answers, repairing 5 and
breaking 9. On LoCoMo an auditor changed 22 answers, improving 2 and breaking 6; three of the six
replaced a correct answer with a hedge. The Terminal-Bench auditors and reviewers of finished
work arose only in ablated loops, and none was admitted. An auditor restricted to counting and state
questions did better (6 repaired, 4 broken), and the self-verification at the memory write later
superseded it. Appendix~\ref{app:verification} gives the full development record.

\begin{table}[t]
\caption{Self-verification admitted in the tested harnesses, and auditors not kept at the final answer.
Development rows use the search-time task models (Appendix~\ref{app:verification}).}
\label{tab:verification-full}
\centering
\scriptsize
\setlength{\tabcolsep}{2pt}
\begin{tabular}{@{}>{\raggedright\arraybackslash}p{0.11\linewidth}>{\raggedright\arraybackslash}p{0.20\linewidth}>{\raggedright\arraybackslash}p{0.13\linewidth}>{\raggedright\arraybackslash}p{0.27\linewidth}>{\raggedright\arraybackslash}p{0.24\linewidth}@{}}
\toprule
Setting & Self-verification & Acts at & Development & Test \\
\midrule
\multicolumn{5}{@{}l}{\emph{Admitted}} \\
AMA-Bench & verify-on-build & memory write & repaired 17 of 18 notes examined & -- \\
LoCoMo & claim judge & evidence merge & best of three placements (0.796 vs 0.763--0.770) & ran on 12--13\% of questions, edited claims on 4--5\% \\
UltraHorizon & commit gate & rule commitment & in the bundle raising validation from 26.7 to 33.3 & switched off: grid 70.8 to 64.1 (Table~\ref{tab:verification-ablation}) \\
Terminal-Bench & edit gate & first edit of a shipped file & repair procedure already wrote both first & not measured \\
\midrule
\multicolumn{5}{@{}l}{\emph{Not kept at the final answer}} \\
AMA-Bench & re-check every answer against raw files & final answer & rewrote 82 of 120; 5 repaired, 9 broken & -- \\
LoCoMo & auditor on every answer & final answer & changed 22; 2 improved, 6 regressed & -- \\
Terminal-Bench & auditors and reviewers of finished work & finished work & none admitted (Table~\ref{tab:ablation}, lower rows) & -- \\
\bottomrule
\end{tabular}
\end{table}

\section{Conclusion}
\label{sec:limitations}

Vestrum improves harnesses it does not own, component by component. It learns from one
to ten long development episodes. Each mechanism is placed where an error arose, aimed at a failure that
recurs within its task kind, and checked by the generalization filter. It raised held-out performance on a closed coding
agent and on a published memory system in all five settings.

\paragraph{Limitations.}
Each setting has one search, so test repetitions measure execution and judging variability only. Every harness is tested inside the benchmark it was
learned on, not transferred to another of the same kind, such as LongMemEval \citep{longmemeval2025} for LoCoMo. GEPA is the only learned baseline, and it has a different editable surface
and a different task model during search. Self-verification was ablated at test on UltraHorizon grid episodes and on
LoCoMo (Appendix~\ref{app:verification-ablation}). The filter tests what a mechanism states, not whether it suits one benchmark's task kinds. PathBench's test sections come from
the same slide as its development sections. LoCoMo excludes unanswerable questions while the harness contains a
best-supported-answer rule, so the LoCoMo gains concern the answerable categories. The next
experiments are repeated searches per setting, learned harnesses transferred across benchmarks, a
test-set run with the filter switched off, and a repeated-sampling baseline at matched budget.

\section*{Ethics Statement}

We evaluate agents on existing benchmarks and spatial transcriptomics measurements. We use
Terminal-Bench~4 Hard, LoCoMo \citep{locomo2024}, AMA-Bench \citep{zhao2026amabench}, and
UltraHorizon \citep{luo2025ultrahorizon} under their released licenses. LoCoMo contains synthetic
dialogues, and UltraHorizon uses simulated environments.

For PathBench, we use publicly available CosMx measurements from the HNSC slide in the AnchorR study
\citep{morris2026anchorr}, comprising six tissue sections. The expert reference annotation comes
from a bioinformatician's analysis pipeline. Records used for evaluation are keyed by section and
cell index, and they contain no patient identifier. The original data are available through the
AnchorR preprint.

\section*{Reproducibility Statement}

Evaluation splits, judging and validation protocols, and detailed results are specified in the
appendices, and Appendix~\ref{app:implementation} gives baseline implementation details. PathBench
uses data publicly available through the AnchorR preprint \citep{morris2026anchorr}.

\section*{AI Use Statement}

AI agents are part of the method studied in this paper. AI also assisted with the literature
survey, idea generation and proofreading.

\bibliography{references}
\bibliographystyle{iclr2027_conference}

\appendix
\section{Comparison with the Closest Harness Searches}
\label{app:related}

Table~\ref{tab:neighbours} sets Vestrum beside the five harness searches closest to it
(Section~\ref{sec:related}).

\begin{table}[h]
\caption{Vestrum beside the closest harness searches
\citep{chen2026harnessfix,jiang2026harnessevolve,zhang2026harnesscompass,wu2026modularrsi,xia2026rrsi}.}
\label{tab:neighbours}
\label{app:neighbours}
\centering
\scriptsize
\setlength{\tabcolsep}{2.5pt}
\begin{tabular}{@{}>{\raggedright}p{0.10\linewidth}>{\raggedright}p{0.13\linewidth}>{\raggedright}p{0.13\linewidth}>{\raggedright}p{0.13\linewidth}>{\raggedright}p{0.13\linewidth}>{\raggedright}p{0.12\linewidth}>{\raggedright\arraybackslash}p{0.14\linewidth}@{}}
\toprule
 & HarnessFix & HarnessEvolve & HarnessCompass & ModularRSI & RRSI & Vestrum \\
\midrule
tasks searched & 34--100 per benchmark & mini-batched training set & 50 & 120 per set & 61--120 & 1 conversation to 10 episodes \\
placement & flaw mapped to its harness layer; localization scored against human labels & first divergence from a reference trajectory & capability edits to code, guidance to prompt or memory & edit restricted to one module & edits tagged by component, memory included & mechanism at the component where the error began \\
task conditioning & repairs record the conditions where they apply & -- & lessons carry an applicability condition & LLM activates functions per task & -- & hard gate to a pre-assigned task kind \\
content filter & spec forbids task data; conformance check & LLM leakage gate; a named contribution & gate in the proposer's prompt; first principle & self-review of the diff, one of three gates & LLM critic before evaluation & separate LLM reviewer and lexical lint, before execution \\
filter evidence & -- & ablation, one dataset & gated vs.\ ungated runs & ablated with other gates & 38 screened, 1 rejected & reviewer refused 5 of 19 on PathBench; removed only with failure classes and scope \\
\bottomrule
\end{tabular}
\end{table}

\section{Evaluation Settings}
\label{app:implementation}

Table~\ref{tab:settings} summarizes the five settings, their data splits, and the units
graded by each benchmark.

\begin{table}[h]
\caption{Evaluation settings. A check is the unit the benchmark grades. Development tasks are the
ones the loop ran and examined. Validation tasks are disjoint from both development and test.
Test tasks were run only after the harness was frozen. At test, GAM runs on
Gemini 3.7 Flash, and Claude Code on Sonnet 5 (Terminal-Bench) or Opus 5 (PathBench).}
\label{tab:settings}
\centering
\small
\setlength{\tabcolsep}{3.5pt}
\begin{tabular}{@{}llllll@{}}
\toprule
setting & check & base harness & development & validation & test \\
\midrule
Terminal-Bench & test & Claude Code & 7 tasks & 4 tasks & 8 tasks $\times$ 2 draws \\
PathBench & cell label & Claude Code & 2 sections & -- & 4 sections $\times$ 2 episodes \\
UltraHorizon & episode score & GAM & 6 episodes & 21 episodes & 42 episodes $\times$ 3 seeds \\
LoCoMo & judged probe & GAM & 1 conversation & 1 conversation & 956 probes $\times$ 3 \\
AMA-Bench & judged probe & GAM & 10 episodes & -- & 240 probes \\
\bottomrule
\end{tabular}
\end{table}

Table~\ref{tab:implementation} gives, for each setting, the components the loop may change, the
task kinds, the failure evidence and the admission criteria. Appendix~\ref{app:loop} describes
the reported Terminal-Bench epoch and quotes its main prompts.

\begin{table}[h]
\caption{How the loop was run in each setting.}
\label{tab:implementation}
\centering
\scriptsize
\setlength{\tabcolsep}{3pt}
\renewcommand{\arraystretch}{1.15}
\begin{tabular}{@{}>{\raggedright\arraybackslash}p{1.8cm}>{\raggedright\arraybackslash}p{2.2cm}>{\raggedright\arraybackslash}p{2.2cm}>{\raggedright\arraybackslash}p{2.2cm}>{\raggedright\arraybackslash}p{2.2cm}>{\raggedright\arraybackslash}p{2.3cm}@{}}
\toprule
& UltraHorizon & LoCoMo & AMA-Bench & PathBench & Terminal-Bench~4 Hard \\
\midrule
components the loop may change &
mechanism settings in the four design dimensions &
question-answering path and synthesis prompts; the store is unchanged &
per-chunk extraction, notebook builds, deterministic views, synthesis prompts &
what the agent reads before it plans &
prompts, instruction files and mechanisms gated by task kind \\
\addlinespace
task kind &
environment &
question category &
probe axis &
one kind, annotation &
task mode $\times$ check style, tagged from the 66 instructions alone \\
\addlinespace
failure classes from &
graded development episodes &
judged failures on the development conversation &
judged probes on 10 development episodes &
per-label disagreement with the reference on the development sections &
failures on the 7 development tasks \\
\addlinespace
development criterion &
each mechanism's predicted behaviour appears on three of the six development episodes, with no collapse &
blinded judgement of paired runs on the development conversation (152 probes) &
paired per-episode differences on judged batches &
paired by seed against the current harness &
gains checks the current harness fails in every draw; no task of another kind loses a check;
cost under $1.5\times$ the baseline \\
\addlinespace
generalization filter &
reviewer; leakage lint &
reviewer; leakage lint &
reviewer; leakage lint &
reviewer; leakage lint &
reviewer; leakage lint \\
\addlinespace
validation set &
21 episodes on new seeds, each used once &
one validation conversation (152 probes), opened during development &
none &
none &
validation tasks, disjoint from the test set \\
\addlinespace
epochs &
four epochs &
three epochs &
four epochs &
one epoch, one iteration &
three epochs; the third is reported, \$123.5 \\
\bottomrule
\end{tabular}
\end{table}

\paragraph{Baseline implementations.}
Apart from the described adapter code, the vendored GAM implementation \citep{gam2025} is unchanged.
A-Mem \citep{amem2025} uses Gemini embeddings, the shared answer prompt, and repaired ingestion.
Claude Code runs on Sonnet~5 for Terminal-Bench and on Opus~5 for PathBench.

\section{Terminal-Bench Ablation: Variant Runs}
\label{app:earlier-loops}

Table~\ref{tab:ablation} in Section~\ref{sec:ablation} ablates the loop of Section~\ref{sec:tb4} by removing parts of the loop.
The second row runs the third epoch without the lessons file. The remaining rows remove parts
cumulatively, in this order: the ceiling probe; failure classes together with the generalization filter and
task-kind scope; turn-level localization; and trace access. Without trace access, the loop sees only final outputs.
Every variant starts from Claude Code and proposes its own mechanisms. We did not ablate the
four design dimensions individually.
In rows that report check counts, a gain is a check that Claude Code fails in every
draw and the evaluated harness passes in every draw. A loss is the reverse. The variant
without localization instead reports the percentage-point improvement in the mean share of checks
passed, from single draws. Cost is per-task spend relative to Claude Code. Only the full loop's frozen harness was run on
the test set.

\paragraph{The lessons file.} Without the lessons recorded after the first two epochs, and with
every other part of the loop in place, the third epoch gained 10 development checks with no losses,
at $1.6\times$ the baseline's cost, above the $1.5\times$ admission limit. With the lessons, it gained
16 development checks against one loss, at $1.32\times$ cost.

\paragraph{The ceiling probe.} The ceiling probe (Appendix~\ref{app:loop}) gives the agent each
development task's solved procedure. It identifies development failures that respond to that
procedure. Without it, the development
gain fell from 16 checks to 3. With the probe, the admitted bundle predicted 12 passing checks,
and 9 of them passed.

\paragraph{The generalization filter and scope.} The variant without failure classes,
the generalization filter and task-kind scope gained 2 development checks at $1.7\times$ cost.
On a task of another kind, the first mechanism this variant tried lost two checks that Claude
Code always passed. That task's score fell from 7 to 5 of 10 checks. Three checks
gained on one development task in one draw did not reproduce in the next. With the
generalization filter and scope, the loop still met a regression from an unscoped mechanism. The
admission criteria refused that mechanism, and its scoped version was then admitted
(Section~\ref{sec:admission}). That loop ran at $0.86\times$ the baseline's cost.

\paragraph{Recognizable failure classes.}
This variant used three fixed classes in place of predicate failure classes. Every failure
fell into the same one. The labels therefore did little to separate one mechanism from another,
and a later iteration found a gain that had been attributed to the wrong mechanism. Failure classes in the
full loop are predicates that can be recognized in traces. Each one specifies behavior for the
proposer to target and for subsequent trace analysis to inspect.

\paragraph{Localization and trace access.}
Without turn-level localization, the loop routed traces into failure clusters, which left 55\%
of failures unassigned. When its second round produced no diagnosis, proposals bypassed
the router. This variant raised its mean share of checks passed from 0.734 to 0.827, at 2.3
times the baseline's cost. Without trace access, the loop read only final
outputs. Its second round changed harness code and role prompts together. The run report
could not determine which change had acted. A repair round lowered a task of another kind
from 17 to 14 of 20 checks. Nothing was admitted.

\section{The Vestrum Loop on Terminal-Bench~4 Hard: Steps and Main Prompts}
\label{app:loop}

We describe the reported Terminal-Bench epoch in the order of
Algorithm~\ref{alg:vestrum} and quote the main prompts, shortened where marked.
The quoted prompts use their original terminology.
Loop calls used Claude Fable~5.1 and Claude Opus~5. Task executions used Sonnet~5,
the same model as the Claude Code baseline.
Every proposal call also reads a fixed design document (\texttt{PRINCIPLES.md}), which the loop never edits.
It states the rules for each design dimension, scoping by task kind and the cost cap.
The reported harness comes from the loop's third epoch. The first two epochs did not improve the
development set; the lessons the loop recorded after them went into the lessons file, which the
third epoch's proposer read. The third epoch cost \$123.5, including its test runs, and the first
two cost about the same. Development used seven tasks, validation used four separate tasks, and
testing used eight further tasks.

\paragraph{Select consistently failed checks.}
Claude Code runs each development task three times. A check contributes failure evidence
only if it fails in all three draws. Intermittent failures, called coin checks in the
original prompts, are excluded from this count but remain available for diagnosis.

\paragraph{Localize and classify.}
One call per failed task reads its Claude Code traces, writes the procedure required to
pass every check, and identifies the missing step for each failure:
\begin{quote}\small
``\#\# Solved trace --- the ordered checklist an agent would have to execute to pass EVERY check,
as observable steps \ldots\ Never quote test code or expected values \ldots\ \#\# Gap --- for each
stable-red AND coin check: which solved-trace steps the bare draws lacked, as CAPABILITIES (KNOW
$\cdot$ TEST $\cdot$ CORRECT $\cdot$ ORDER \ldots).''
\end{quote}

\paragraph{Ceiling probe (Terminal-Bench only).} Before any mechanism was proposed, each
development task was run once with its solved procedure given to the agent. On five of the seven
tasks, this run passed checks that Claude Code never passes. The probe thus identified failures
that respond to procedural assistance. On the other two it did not, and proposals targeted only
the five responsive tasks.

\paragraph{Compose and filter.}
One composer call proposes, selects, and composes the bundle. The generalization filter
therefore runs after selection and before any candidate is executed. From the missing steps, the
composer produces mechanisms scoped to the task kind:
\begin{quote}\small
``Your job is to make the harness DELIVER the DELIVERY steps to an agent that has only the
request, on every task of the same shape, without naming the task. Start from the parent: keep
every parent component that ever passed a stable-red check \ldots\ Then add 2--4 new
components.''
\end{quote}
Each mechanism then goes through the generalization filter. A reviewer answers seven pass/fail
questions per mechanism, and the leakage lint refuses any string or constant unique to one task.
\begin{quote}\small
``Q1 fails if the decision is a fixed vocabulary/regex list \ldots\ Q4 fails if anything
task-derived is baked in (check the code, not the claim).''
\end{quote}

\paragraph{Evaluate, prune, and reevaluate.}
The bundle is evaluated against the current harness on paired development tasks, and a
trace reader labels each mechanism's behavior:
\begin{quote}\small
```not\_fired' \ldots\ `fired\_not\_acted' (delivered or ran, but the agent's actions at the
targeted point do not differ from a bare run), `fired\_acted' (the agent visibly did what the
component asked \ldots)''
\end{quote}
Mechanisms judged inactive are disabled and those judged harmful are dropped before the
pruned bundle is reevaluated:
\begin{quote}\small
``a label is WORKED only if the stable-red check passed AND the read shows the component's own
artifact/turn caused the behaviour \ldots\ FIX if NOT-FIRED \ldots\ DROP if HARMFUL \ldots\ add
new components ONLY along the lines of what worked.''
\end{quote}
In the reported epoch, the full bundle gained 16 checks that Claude Code failed in every
draw and lost one that it passed in every draw. Its proposals predicted 12 passing checks, and
9 of them passed. The pruned bundle gained 11 and lost 4, so
only the full bundle advanced to validation. Development runs cost 1.32 times the baseline.

\paragraph{Admission criteria.} A bundle is admitted when it passes checks that Claude Code fails
in every draw, loses no check that the current harness passed on a task of another kind, and
costs under $1.5\times$ the Claude Code mean per task (Section~\ref{sec:admission}).
Development results determine which of the full and pruned bundles advance to validation. A
pruned bundle is validated only if it outperforms the full bundle on development. Final
admission follows validation.

\paragraph{Validation and lessons.} Before the harness is admitted and frozen, it runs on the
validation tasks to confirm the development result. These tasks are disjoint from the test set.
Each epoch closes with a meta-review call that names every line of proposals the epoch tried,
records its outcome and writes a task-free lesson to the lessons file. A line that fails twice
goes to a dead list, which later compositions may not re-propose.

\paragraph{Test.} The frozen admitted harness is evaluated on the eight test tasks, with two
draws per system. Test outcomes are not returned to search.

Results are in Table~\ref{tab:tb4-test8}. Under the same per-task spend cap, the frozen harness's
mean test cost per task was $1.03\times$ Claude Code's. This includes the request classifier, a
Claude Haiku~4.5 call, and its lookups of standard definitions.

\section{The Generalization Filter and Permissible Supervision}
\label{app:population}

\paragraph{What the filter rejects.}
The harness may contain a procedure, keyed on terms that appear in the request. It may not
contain a fact taken from a development task. On Terminal-Bench, for example, the procedure
that obtains a file-format convention or an estimator's precondition may be written into the
harness, but an answer value from a development task may not. A knowledge class found in no
other task is flagged. A mechanism that depends on it is treated as overfitted to one task. The
leakage lint searches a mechanism's text for task-specific names, symbols, paths and constants.
A language-model reviewer reads the mechanism's condition and content (the Terminal-Bench
questions are quoted in Appendix~\ref{app:loop}). Admission and held-out evaluation follow the
generalization filter.

\paragraph{Task kinds on Terminal-Bench~4 Hard.} A separate pass reads each of the 66
instructions once, with no access to solutions or tests. It records each instruction's mode
(repair, build, configure or analyze) and the style of its checks (a
specification file, an oracle, or measured values). A task's kind is its mode crossed with its
check style. The public instruction list that the pass reads includes test instructions. For a
test task, a proposal sees only the count of tasks of its kind. It never sees the task's
solution, test, trace or outcome.

\subsection{Permissible Supervision by Setting}
\label{app:supervision}

In all five settings, a mechanism may contain a procedure, but no task-specific name, symbol,
path or constant from an instruction or a test. The settings differ in what
else is permitted.

\begin{itemize}
  \item \textbf{UltraHorizon, AMA-Bench.} Nothing taken from a development episode's board,
  rules or graded answer.
  \item \textbf{LoCoMo.} Two answer rules written from development failures. Nothing taken from
  a development probe's answer.
  \item \textbf{PathBench.} Rules, reference tables and a helper written from the failures and
  reference labels of the two development sections. Nothing is taken from the four test sections.
  \item \textbf{Terminal-Bench.} Nothing taken from the solution, test, trace or outcome of a test
  task. Of its instruction, only its task kind.
\end{itemize}

\section{Validation Protocol on UltraHorizon}
\label{app:gates}

On UltraHorizon, a draw is one complete long-horizon episode. Development episodes screen proposals,
and validation decides admission. A mechanism passes this development screen if its predicted
behavior appears on three of the six development episodes and no score collapses relative to the
current harness. A commit gate that holds back an answer is one such behavior.

For validation, new seeds are sampled outside every manifest, and both the candidate and the current
harness run on each seed. Each seed is used once. Only the decision and the mean score return to the
loop. No transcript or per-episode score returns. Over four epochs and 21 validation seeds, the loop
considered 54 proposals and admitted two bundles, which together contain the UltraHorizon mechanisms
in Table~\ref{tab:components}.

\section{Memory Benchmark Setups}
\label{app:memory-setup}

The benchmarks test different uses of retained information. In UltraHorizon, the agent must discover
environment rules through long interactions \citep{luo2025ultrahorizon}. Development uses six
grid episodes, three of which screen proposals. Each of the three test seeds uses 42 episodes: 21 grid,
14 sequence and 7 biology. The sequence and biology environments were left out of development by
accident, so they form an out-of-development test. The mechanisms learned on grid episodes therefore
also run in the sequence and biology environments at test. LoCoMo poses
questions about extended conversations \citep{locomo2024}. We evaluate its multi-hop, temporal,
open-domain, and single-hop questions and exclude the adversarial, unanswerable category. Answers
are scored by strict accuracy, which requires all key facts without contradiction. We report
lexical F1 as a secondary measure, given its known limitations \citep{li2026locomoplus}. Development
uses one conversation of 152 probes, and validation uses a second. Testing uses six other
conversations with 956 probes, run in three replicates with fresh stores and judging runs. AMA-Bench tests
recall, causality, state updating, and state abstraction after tool-using trajectories
\citep{zhao2026amabench}. It has ten development episodes.
Its twenty held-out episodes contain 240 probes.

\section{Judge Protocol}
\label{app:judge}

On the memory benchmarks, a blinded Sonnet judge grades each answer against a fixed rubric in three
independent passes, and the majority vote decides. The judge compares the answer with the reference
key facts without learning which system produced it. Judge accuracy counts an answer only when all
required facts are present and there is no contradiction. The detailed tables call this measure
strict accuracy. Lenient accuracy also counts partial answers. Across the three LoCoMo runs, pass agreement
was 0.955, 0.947, and 0.948. On AMA-Bench, 663 of 720 judgments were unanimous.

For UltraHorizon, grading uses the benchmark rubric for each environment and one canonical judge
prompt. Every environment is graded on a 0--100 scale, and both systems are judged in one 42-episode
batch per seed. The same rule library recurs across episodes. The statistical units are therefore
the rule configurations and not the 126 episode executions. For that reason, the main text reports
paired directions and seed means without an episode-level $p$-value.

\section{Full Result Tables}

\subsection{UltraHorizon: full seed record}
\label{app:uh-seeds}

Each seed contains the same 42 test episodes per system. Table~\ref{tab:uh-seeds} records the
mean of every seed.

\begin{table}[h]
\caption{UltraHorizon mean score per 100 over the 42 test episodes of each seed.}
\label{tab:uh-seeds}
\centering
\small
\begin{tabular}{c ccc}
\toprule
Seed & GAM & Vestrum & $\Delta$ \\
\midrule
1 & 47.86 & \textbf{54.88} & $+7.02$ \\
2 & 48.10 & \textbf{60.12} & $+12.02$ \\
3 & 46.79 & \textbf{64.29} & $+17.50$ \\
\midrule
pooled & 47.58 & \textbf{59.76} & $+12.18$ \\
\bottomrule
\end{tabular}
\end{table}

\subsection{LoCoMo: full seed record}
\label{app:locomo-seeds}

\begin{table}[h]
\caption{LoCoMo strict and lenient accuracy by full run, 956 test probes per system.
Net is Vestrum's probe wins minus losses against GAM; $p$ is an exact sign test over probes.
Agreement is the fraction of
grades unanimous across the three judge passes, computed over the whole replicate and therefore
the same for every system in it.}
\label{tab:locomo-seeds}
\centering
\small
\begin{tabular}{c l ccccrr}
\toprule
Replicate & System & Strict & Lenient & Lexical F1 & Agreement & Net & $p$ \\
\midrule
1 & Vestrum & 0.741 & 0.881 & 0.667 & 0.955 & $+24$ & 0.038 \\
1 & GAM      & 0.715 & 0.859 & 0.618 & 0.955 & --- & --- \\
1 & A-Mem    & 0.497 & 0.647 & 0.477 & 0.955 & --- & --- \\
2 & Vestrum & 0.749 & 0.897 & 0.679 & 0.947 & $+39$ & $5.6\!\times\!10^{-4}$ \\
2 & GAM      & 0.708 & 0.865 & 0.612 & 0.947 & --- & --- \\
2 & A-Mem    & 0.497 & 0.651 & 0.480 & 0.947 & --- & --- \\
3 & Vestrum & 0.756 & 0.901 & 0.679 & 0.948 & $+34$ & $2.9\!\times\!10^{-3}$ \\
3 & GAM      & 0.721 & 0.865 & 0.623 & 0.948 & --- & --- \\
3 & A-Mem    & 0.486 & 0.632 & 0.473 & 0.948 & --- & --- \\
\bottomrule
\end{tabular}
\end{table}

On LoCoMo, the statistical unit is the conversation. Questions within a conversation share retrieved
evidence, so probe-level $p$ values are optimistic. The main text therefore reports the direction of each
replicate. Per run, GAM used 4,427/4,479/4,468 calls at \$15.39/\$15.51/\$15.55. Vestrum used
4,104/4,192/4,130 at \$21.20/\$21.73/\$21.34, and A-Mem used 12,029/12,040/12,025 at
\$17.63/\$17.49/\$17.46. Vestrum's planner split 12 of 152 questions on the development conversation
and 11 of 152 on the validation conversation.

\subsection{AMA-Bench: full three-way and replication record}
\label{app:ama-seeds}

\begin{table}[h]
\caption{AMA-Bench held-out three-way result, 240 shared probes under one blinded three-pass
judgment. Vestrum and GAM use their second answer-generation run; A-Mem has one generation.
Per-axis counts are StateAbstraction 40, StateUpdating 67, Causal 58 and Recall 75.}
\label{tab:ama-new}
\centering
\small
\begin{tabular}{lccccc}
\toprule
System & Overall & StateAbs. & StateUpd. & Causal & Recall \\
\midrule
Vestrum & \textbf{0.742} & \textbf{0.78} & \textbf{0.78} & \textbf{0.76} & \textbf{0.68} \\
GAM \citep{gam2025} & 0.642 & 0.57 & 0.75 & 0.67 & 0.56 \\
A-Mem \citep{amem2025} & 0.562 & 0.45 & 0.57 & 0.60 & 0.59 \\
\bottomrule
\end{tabular}
\end{table}

\begin{table}[h]
\caption{AMA-Bench Vestrum--GAM replication. Each row is a within-run comparison on the same 20
episodes and 240 probes. Every $p$ is a two-sided exact sign test over episodes.}
\label{tab:ama-replication}
\centering
\small
\begin{tabular}{c ccccr}
\toprule
Run & Vestrum & GAM & $\Delta$ & Episode W/L/T & sign-test $p$ \\
\midrule
1 & 0.787 & 0.667 & $+0.121$ & 12/3/5 & 0.0352 \\
2 & 0.742 & 0.642 & $+0.100$ & 11/4/5 & 0.118 \\
\bottomrule
\end{tabular}
\end{table}

To measure generation and judging variability, we evaluated the frozen harness twice. The three-way comparison
uses the second run, which appears in Table~\ref{tab:ama-new} and in the upper
block of Table~\ref{tab:memory}. Both paired run comparisons are recorded in
Table~\ref{tab:ama-replication}. In the final three-way run, Vestrum's margin over A-Mem is $+0.179$
strict accuracy, with 13/4/3 episode wins, losses and ties ($p=0.0490$ two-sided).

\paragraph{Test cost.}
UltraHorizon seeds 2 and 3 used 5,053 and 4,872 model calls (\$49.30 and \$45.90). The AMA-Bench
Vestrum runs used 713 and 724 calls. LoCoMo used 61,894 calls and \$163.30 across all systems
and seeds.

\section{PathBench Grading}
\label{app:pathbench}

In PathBench, an agent must recover tissue sections from a spatial transcriptomics export, filter
cells, cluster their expression profiles, and assign names from sixteen cell types. The agent has
seven published sources, two in-house reference files and no network. The reference is the laboratory annotation that a
bioinformatician's pipeline produced for the AnchorR data \citep{morris2026anchorr}. Development
uses two sections of a head-and-neck squamous cell carcinoma slide, with 55,366 reference cells.
Testing uses the other four sections, with about 129,000 cells scored. Before the
harness was frozen, no test-section label or measurement was read.

The grader (\texttt{score\_run.R} and \texttt{partition.py}) joins agent labels to the laboratory
reference exactly on cell identifier, within the sections of the split. The join matches 99.0 to
99.9 percent of cells in the agent's output. The four test sections contain 136,338 reference cells,
and about 129,000 of them remain after the agent's quality filter and are scored. Reference cells
the agent's output does not contain are reported separately as coverage.

The agent must use the sixteen labels the task brief gives, and any other label never matches.
Reference cells labelled Squam3, a class with no label in the brief, are removed before any metric.
Accuracy is the share of scored cells whose label matches the reference. Macro-F1 is the unweighted
mean F1 over the fifteen classes other than Unknown that occur in the reference. The endpoint is
agreement with an expert-in-the-loop laboratory annotation \citep{morris2026anchorr}. The underlying data are publicly available through the
AnchorR preprint.

The adjusted Rand index (ARI) and normalized mutual information (NMI) compare the agent's grouping
of cells with the reference grouping, independently of label names. Accuracy and macro-F1 compare cell-type assignments. We also count the section--class pairs with F1 $\geq 0.5$. On development,
Claude Code and Vestrum average 18.0 and 27.5 of 30 pairs. On test, they average 43.5 and 51.5 of 60
pairs.

\section{GEPA Comparison and Resource Accounting}
\label{app:gepa}

\paragraph{Search configuration.}
We ran GEPA \citep{agrawal2025gepa} version 0.1.4 with stock reflective evolution. It optimized GAM's
question-time prompts on LoCoMo and AMA-Bench and its action protocol on UltraHorizon,
with memory construction fixed. Its task model was Gemini~3.7 Flash, as for every system
at test, and its reflection model was Claude Opus~5. Vestrum searched using the earlier task
models specified in Section~\ref{sec:experiments}. Evaluation counts therefore compare
numbers of task-agent executions, not executions of the same model. GEPA received
Vestrum's development data and used its own Pareto set for selection. On UltraHorizon,
GEPA's reflection received the judge's per-rule verdicts but never the ground-truth rules.
Every GEPA candidate passed Vestrum's leakage lint. An earlier run whose candidate embedded
the answer key in its prompt was discarded.

\paragraph{Test and judging repetitions.}
The frozen GEPA program was run on three test seeds. On LoCoMo, each seed was judged
blind together with GAM, Vestrum, and A-Mem. AMA-Bench used three judging runs: the third
paired GEPA's third generation with GAM's and Vestrum's first-generation answers.
UltraHorizon used the canonical judge for each seed. As in Appendix~\ref{app:judge},
each answer is graded against the reference answer's key facts. Judge accuracy is the
share graded fully correct.

These evaluations form a separate comparison from the upper block of
Table~\ref{tab:memory}. That block reports UltraHorizon means over three seeds
(126 paired executions), LoCoMo mean and standard deviation over three replicates, and
AMA-Bench's second answer-generation run (240 probes). Its A-Mem row is one generation using
Gemini embeddings. The lower block reports the separately judged GEPA comparison.

\paragraph{Performance.}
Vestrum scores above GEPA in every reported paired evaluation
(Table~\ref{tab:gepa-within}). On AMA-Bench, GEPA improves over GAM by a pooled net
$+37$ questions. Vestrum's further margin is concentrated in state abstraction. Vestrum's
trajectory views serve these questions, and GEPA's prompts do not reach them. On UltraHorizon, GEPA completed four
iterations without improving GAM's prompt on its Pareto set and stopped at a plateau.
GEPA's test results therefore re-execute GAM's program. The 44 wins, 38 losses, and 44 ties
against GAM across 126 episodes describe execution variability.

\begin{table}[h]
\caption{GEPA against GAM and Vestrum over three test seeds (AMA-Bench: three judging runs). Judge
accuracy on LoCoMo and AMA-Bench, score per 100 on UltraHorizon. Net counts questions or episodes
won minus lost, pooled over seeds, ties excluded; $p$ is a two-sided exact sign test over questions,
optimistic because questions in one conversation or episode share evidence.}
\label{tab:gepa-within}
\centering
\footnotesize
\setlength{\tabcolsep}{3pt}
\begin{tabular}{@{}>{\raggedright\arraybackslash}p{0.20\linewidth}>{\raggedright\arraybackslash}p{0.17\linewidth}>{\raggedright\arraybackslash}p{0.17\linewidth}>{\raggedright\arraybackslash}p{0.17\linewidth}>{\raggedright\arraybackslash}p{0.23\linewidth}@{}}
\toprule
 & GAM & GEPA & Vestrum & Vestrum $-$ GEPA \\
\midrule
LoCoMo, seeds 1/2/3 & 72.9 / 71.7 / 72.9 & 74.0 / 73.2 / 74.7 & \textbf{75.5 / 76.6 / 76.7} & net $+15$ / $+32$ / $+19$ \\
\quad mean $\pm$ s.d. & $72.5 \pm 0.7$ & $74.0 \pm 0.8$ & $\mathbf{76.3 \pm 0.7}$ & pooled net $+66$, $p = 5.7\times10^{-4}$ \\
AMA-Bench, runs 1/2/3 & 65.0 / 66.7 / 65.4 & 72.5 / 70.4 / 69.6 & \textbf{77.1 / 75.4 / 76.7} & net $+11$ / $+12$ / $+17$ \\
\quad mean $\pm$ s.d. & $65.7 \pm 0.9$ & $70.8 \pm 1.5$ & $\mathbf{76.4 \pm 0.9}$ & pooled net $+40$, $p = 6.3\times10^{-4}$ \\
UltraHorizon, seeds 1/2/3 & 47.9 / 48.1 / 46.8 & 47.1 / 46.4 / 50.1 & \textbf{54.9 / 60.1 / 64.3} & $+7.8$ / $+13.7$ / $+14.2$ \\
\quad mean $\pm$ s.d. & $47.6 \pm 0.7$ & $47.9 \pm 2.0$ & $\mathbf{59.8 \pm 4.7}$ & 68 / 26 / 32 episodes (W/L/T) \\
\bottomrule
\end{tabular}
\end{table}

\paragraph{Learning tokens and stopping conditions.}
Table~\ref{tab:gepa-budget} distinguishes budget caps from actual usage. On LoCoMo, GEPA's
cap was 78.2M tokens, but it stopped at a plateau after 22.8M. On AMA-Bench its cap was
raised to 25M, and it used 15.7M before reaching its time limit, compared with Vestrum's
9.5M. On UltraHorizon it also ended at a plateau. Token counts sum usage over all models
called by each loop. Vestrum uses
more learning tokens on LoCoMo and UltraHorizon and fewer on AMA-Bench.

\begin{table}[h]
\caption{Learning budget. Tokens are summed over every model each loop called.}
\label{tab:gepa-budget}
\centering
\small
\begin{tabular}{lccccc}
\toprule
 & Vestrum tokens & GEPA cap & GEPA used & GEPA reflection calls & GEPA stopped at \\
\midrule
LoCoMo & $\sim$30M & 78.2M & 22.8M & 10 & plateau \\
AMA-Bench & 9.5M & 25M & 15.7M & 8 & time limit \\
UltraHorizon & $\sim$74M & -- & 21.9M & 4 & plateau \\
\bottomrule
\end{tabular}
\end{table}

\paragraph{Estimated task-agent evaluations.}
Task-agent evaluations measure a different resource: executions of the program being
improved. GEPA evaluates the current and proposed candidates on batches of 32 development
probes, or six UltraHorizon episodes, and reevaluates improving candidates on its full
validation set. Vestrum evaluates composed bundles, reuses prior current-harness runs where
available, and reads existing traces. Its totals use an allowance of 1.5 passes per epoch
for pruning and retesting. A pruned bundle reaches validation only if it
outperforms the full bundle on development. Table~\ref{tab:gepa-calls} gives the calculation.

\begin{table}[h]
\caption{Estimated task-agent evaluations during learning under the stated accounting convention:
probe answers on LoCoMo and AMA-Bench, episodes on UltraHorizon. Vestrum's totals use the
$1.5\times$ allowance for pruning and retesting.}
\label{tab:gepa-calls}
\centering
\small
\begin{tabular}{@{}l>{\raggedright\arraybackslash}p{4.3cm}>{\raggedright\arraybackslash}p{4.3cm}c@{}}
\toprule
 & GEPA & Vestrum & Vestrum / GEPA \\
\midrule
LoCoMo & 64 iterations $\times$ 64, plus 9 validation passes $\times$ 81: \textbf{4{,}825} &
3 epochs $\times$ 1.5 $\times$ (152 development $+$ 152 validation): \textbf{1{,}368} & $0.28\times$ \\
AMA-Bench & 42 iterations $\times$ 64, plus 5 validation passes $\times$ 48: \textbf{2{,}928} &
4 epochs $\times$ 1.5 $\times$ 120 development: \textbf{720} & $0.25\times$ \\
UltraHorizon & 54 batch episodes, plus 6 validation: \textbf{60} &
4 epochs $\times$ 1.5 $\times$ 3 development, plus 21 validation seeds $\times$ 2: \textbf{60} & $1.00\times$ \\
\bottomrule
\end{tabular}
\end{table}

\paragraph{Interpretation and uncertainty.}
Under this accounting, Vestrum uses about a quarter of GEPA's task-agent evaluations on
LoCoMo and AMA-Bench and the same number on UltraHorizon. It scores higher in every paired
evaluation. It spends more learning tokens on LoCoMo and UltraHorizon, so its saving is in
task execution, not in total compute.

The pooled question-level sign tests favor Vestrum on LoCoMo and AMA-Bench
($p = 5.7\times10^{-4}$ and $6.3\times10^{-4}$). The comparison sets a component-wise search against a prompt-only
search on the same development data, each with its own task model during search.

\section{Admitted Mechanisms in Full}
\label{app:components}

Table~\ref{tab:components_full} lists the admitted mechanisms by setting, and
Table~\ref{tab:components} groups them by design dimension.

\begin{table}[p]
\centering
\caption{Admitted mechanisms. Self-verification rows specify the input, the proposition checked, the
evidence and the response on failure. Other rows describe the operation. A dash indicates an unspecified field.}
\label{tab:components_full}
\scriptsize
\setlength{\tabcolsep}{2.5pt}
\renewcommand{\arraystretch}{1.0}
\begin{tabular}{@{}>{\raggedright\arraybackslash}p{0.09\linewidth}>{\raggedright\arraybackslash}p{0.12\linewidth}>{\raggedright\arraybackslash}p{0.13\linewidth}>{\raggedright\arraybackslash}p{0.11\linewidth}>{\raggedright\arraybackslash}p{0.15\linewidth}>{\raggedright\arraybackslash}p{0.115\linewidth}>{\raggedright\arraybackslash}p{0.115\linewidth}@{}}
\toprule
Setting & Mechanism & Dimension & Input & Proposition checked & Evidence & On failure \\
\midrule
Ultra\-Horizon & Commit gate & self-verification & a drafted answer and its rule class & every logged
observation of that class has been replayed against the answer & the pinned per-class notebook &
the answer is held, up to three times \\
& Re-analysis subcall & decompo\-sition & the observation rows of one rule class & the rows of the
class do not contradict each other & the provenance-stamped rows & a nested call re-analyses
that one rule \\
& Per-class notebook & retrieval & \multicolumn{4}{p{0.52\linewidth}@{}}{Observation rows and a
confound matrix, pinned per rule class.} \\
& Probe plans & decompo\-sition & \multicolumn{4}{p{0.52\linewidth}@{}}{The agent plans
deliberate re-probes instead of waiting for evidence to arrive.} \\
& Evidence rows & knowledge synthesis & \multicolumn{4}{p{0.52\linewidth}@{}}{Provenance-stamped, differenced observation rows written as evidence arrives.} \\
\midrule
LoCoMo & Split-or-keep planner & decompo\-sition & \multicolumn{4}{p{0.52\linewidth}@{}}{Decides whether to split
the question; two rules written from development failures keep counterfactual and
single-entity-set questions whole.} \\
& Claim judge at merge & self-verification & the claims of an answer synthesised across sub-queries
& each claim is supported by the spans retrieved for it & the retrieved spans & -- \\
& Re-research round & retrieval & \multicolumn{4}{p{0.52\linewidth}@{}}{One further retrieval
round when the selection for a sub-query is insufficient.} \\
& Two answer rules & knowledge synthesis & \multicolumn{4}{p{0.52\linewidth}@{}}{Order claims by strength
and drop none; commit to a best-supported guess instead of refusing.} \\
\midrule
AMA-Bench & Verify-on-build notebooks & self-verification & a notebook entry as it is integrated &
the entry is supported by the chunk it was extracted from & its extracts, cited raw turns and the action ledger & the note is repaired \\
& Deter\-ministic views & retrieval & \multicolumn{4}{p{0.52\linewidth}@{}}{An event index, an
action record, tallies, object timelines and an entity index, always on and compacted.} \\
& Write-time notebooks & knowledge synthesis & \multicolumn{4}{p{0.52\linewidth}@{}}{Per-notebook
buffers of per-chunk extractions, from which notebooks are integrated as they fill.} \\
\midrule
Path\-Bench & Rule file & knowledge synthesis & \multicolumn{4}{p{0.52\linewidth}@{}}{Eleven rules,
each a statement with its sources, delivered once before the agent plans; for example, every
cluster with a lineage is named by that lineage, and \emph{Unknown} is one specific cluster.} \\
& Reference tables & knowledge synthesis & \multicolumn{4}{p{0.52\linewidth}@{}}{A marker table by label and expression profiles by label and by reference cluster, built from the two development sections.} \\
& Nearest-neighbour helper & knowledge synthesis & \multicolumn{4}{p{0.52\linewidth}@{}}{A script that reassigns the cells of small unlabelled clusters to their nearest labelled neighbours.} \\
\midrule
Terminal-Bench~4 Hard & Edit gate & self-verification & the first change to a shipped file on a repair task &
an invariant and a probe harness exist & the workspace & the change is refused \\
& Repair procedure & knowledge synthesis & \multicolumn{4}{p{0.52\linewidth}@{}}{A procedure for
repair tasks, delivered before the first turn.} \\
& Data-repair procedure & knowledge synthesis & \multicolumn{4}{p{0.52\linewidth}@{}}{A procedure
for repair tasks with shipped data, delivered before the first turn.} \\
& Scale decision & knowledge synthesis & \multicolumn{4}{p{0.52\linewidth}@{}}{A decision with
two fixed branches, delivered on analysis tasks with a measured-values test.} \\
\bottomrule
\end{tabular}
\end{table}

\paragraph{From development failures to mechanisms.}
The following summaries connect each development diagnosis to the corresponding admitted
bundle.

\paragraph{Terminal-Bench 4 Hard.} Claude Code changed shipped code without recording the code's invariants. Before the first change,
the edit gate requires an invariant and a probe script. Two repair procedures specify what to
record, and a scale decision records the chosen branch before execution.

\paragraph{UltraHorizon.}
On the medium and hard development boards, GAM scored zero, and it committed answers without
replaying their supporting observations. The learned harness organizes observations into evidence
rows and a notebook for each rule class. Contradictory evidence is re-analyzed by a separate model
call, and probe plans guide investigation. Where the commit gate is enabled, it replays the logged observations
before accepting a rule (Appendix~\ref{app:verification}).

\paragraph{LoCoMo.} A planner decides whether to split the question. Split questions
receive one further retrieval round when evidence is insufficient. A claim judge then acts where the sub-answers merge.
Two answer rules place the strongest evidence first and require a best guess instead of a refusal.

\paragraph{AMA-Bench.}
Development failures implicated notebook entries used by later probes. The admitted bundle
preserves notebook history, verifies entries during construction, and provides deterministic
views including an event index, an action record, and object timelines.

\paragraph{PathBench.} Claude Code assigned thousands of cells to Unknown although markers indicated their lineage. Three
mechanisms provide guidance before the agent plans: eleven sourced rules, the first of which names
each cluster by its lineage; marker and expression reference tables derived from the development
sections; and a nearest-neighbour helper for small unlabelled clusters.

\section{Self-Verification: The Full Development Record and Activity at Test}
\label{app:verification}

This appendix gives the record behind Section~\ref{sec:verification} and
Table~\ref{tab:verification-full}: every self-verification the searches kept and every alternative they tried,
including the Terminal-Bench ablation variants of Table~\ref{tab:ablation} and earlier harness
versions. Those experiments used GPT-4o-mini on AMA-Bench, Claude Haiku~4.5 on LoCoMo and
UltraHorizon, and Claude Sonnet~5 on Terminal-Bench. Each comparison is paired within its own run
and uses that run's task model.

\begin{table}[h]
\caption{The self-verification record. Stage is where each self-verification acts. Split: ``dev'' is the development
set of that run, ``val'' its validation set. The first row of each setting is the self-verification
in the tested harness; the other rows were not kept.}
\label{tab:verification-record}
\centering
\scriptsize
\setlength{\tabcolsep}{1.5pt}
\begin{tabular}{@{}>{\raggedright\arraybackslash}p{0.11\linewidth}>{\raggedright\arraybackslash}p{0.20\linewidth}>{\raggedright\arraybackslash}p{0.12\linewidth}>{\raggedright\arraybackslash}p{0.19\linewidth}>{\raggedright\arraybackslash}p{0.29\linewidth}>{\raggedright\arraybackslash}p{0.05\linewidth}@{}}
\toprule
Setting & Self-verification & Stage & Grounded in & Result & Split \\
\midrule
AMA-Bench & verify-on-build (kept) & memory write & extracts, cited raw turns, action ledger & repairs the note, never the answer & dev \\
& every answer vs raw files & final answer & raw trajectory files & rewrote 82/120; 5 repaired, 9 broken; strict 0.263 vs 0.297 & dev \\
& answer rewriter & final answer & draft's evidence & 17/48 $\to$ 15/48 (1 rescued, 3 corrupted) & dev \\
& selective, conservative answer auditor & final answer, gated & raw trajectory files & fired 44/120, changed 12; 6 repaired, 4 broken; superseded by design & dev \\
\midrule
LoCoMo & claim judge at merge (kept) & evidence merge & cited retrieved spans & best of three pipelines (Table~\ref{tab:placement}) & dev \\
& post-answer auditor, every answer & final answer & retrieved material & 22 answers changed: 2 improved, 6 regressed; its ``grain'' test alone 2 improved, 0 regressed & dev+\allowbreak 2nd \\
& claim audit widened beyond merges & claims of any answer & retrieved material & 17 answers changed: 0 improved, 3 regressed & dev+\allowbreak 2nd \\
\midrule
Ultra\-Horizon & commit gate (kept) & commit & replay of logged rows & dev 30.0 vs 16.7; val 33.3 vs 26.7 & val \\
& commit gate, first form & write + commit & logged rows & alone, committed 2/6 episodes; clean re-run 36.0 vs 36.0 & dev \\
& rival slate, withheld-row replay & write + commit & withheld rows & val 26.7 vs 53.3 & val \\
& sufficiency gate & commit & distinct values per predicate & held every episode; 26.7 vs 36.7 & dev \\
\midrule
Terminal-Bench & edit gate (kept) & before an edit & hook: invariant and probe exist & admitted within the bundle of four & dev \\
& constraint checks after each call & after each call & deterministic predicates & admitted by an ablated loop & dev \\
& audit checklist; two-phase audit & finished work & separate auditor call & rewards 1/10 and 0/10; two grader-confirmed fixes; refused & dev \\
& tool-using auditor & finished work & executed adversarial cases & one task 13--15/15 vs 11; within noise on six; 3.2$\times$ tokens & dev \\
& adversary on delivered code & finished work & executable probes & two tasks improved; 18--43$\times$ tokens & dev \\
& non-author reviewer & finished work & separate model call & no hidden check moved; new loss; 2.1$\times$ & dev \\
& self-graded check per requirement & finished work & agent's own checks & gains on two tasks, new losses; 1.9$\times$ & dev \\
& second draw and reconcile & finished work & an independent draw & reconcile starved of budget; 2.2--6$\times$ & dev \\
& once-per-session stop gates & at stop & hook over agent's ledger & acted once; moved nothing & dev \\
& pre-fix self-graded probes & before an edit & agent's own probes & identical to Claude Code; 1.3$\times$ & dev \\
& ledger required before first write & before an edit & hook & no gain over Claude Code & dev \\
\bottomrule
\end{tabular}
\end{table}

\paragraph{AMA-Bench.}
The kept self-verification compares each notebook entry with its source extracts, cited raw turns and the
deterministic action ledger. A failed comparison repairs the note, so later questions use the corrected
record. In development, with GPT-4o-mini as the writer, it flagged and repaired 17 of the 18 notes it
checked. The refused auditor re-checked every final answer against the raw
trajectory files, with the instruction that the raw record overrides the draft. It rewrote 82 of
120 development answers, repairing 5 and breaking 9 (turning them wrong or partial). The
development record states ``always-verify second-guesses correct answers''. A gated version of the
same auditor ran only on counting and state questions and was told to keep the draft unless the
raw record contradicts it. This gated auditor fired on 44 questions, changed 12, and was net
positive (6 repaired, 4 broken). It was not refused on evidence: the final design moved
self-verification from the answer to the notebook, and the gated auditor was not evaluated again.

\paragraph{LoCoMo.}
Table~\ref{tab:placement} compares three development pipelines that run span selection, span
checking and claim judging on every answer, on none, or only at sub-query evidence merges. The
merge-only pipeline also includes planner rules derived from the same conversation, so it split 12
questions while the other two split 48 to 53. Placement and routing differ together. Judging was
blind and three-pass. The merge-only pipeline leads each alternative by four answers net. On a second conversation it scored 0.836, against 0.829 for a
routing baseline and 0.809 for GAM. The harness based on it gained over GAM on the test conversations
(Table~\ref{tab:memory}). The kept claim judge acts rarely: on the test conversations it ran on 12
to 13\% of questions, edited claims on 4 to 5\%, and dropped at most four claims per replicate. The
refused answer-level auditors were measured within each question: the same question was answered
from the same evidence with and without the auditor, and both answers were graded blind. The auditor
that ran on every answer changed 22 answers, improving 2 and regressing 6. Three of the six regressions
replaced a correct answer with a hedge such as ``This cannot be confirmed from the material
provided''. One of its tests, on the grain of the answer, was positive on its own (2 improved,
none regressed). The claim audit that was widened beyond the merge points changed 17 answers,
regressing 3 and improving none. At the level of whole runs, the variant with both auditors scored
0.822 against 0.803 for the kept harness, pooled over two conversations. A whole-run score mixes
the auditors with the rest of the pipeline; the within-question comparison measures the auditors alone.

\begin{table}[h]
\caption{LoCoMo placement comparison on the development conversation: three pipelines that differ
in where span selection, span checking and the claim judge run. Strict accuracy by question path;
$n$ is the number of questions on that path. The kept row is the judging run that gave 0.796.}
\label{tab:placement}
\centering
\small
\begin{tabular}{lccc}
\toprule
Checking runs & single path & split path & overall \\
\midrule
on every answer & $\approx$0.79 & 0.736 ($n{=}53$) & 0.763--0.770 \\
on no answer & \textbf{0.856} ($n{=}104$) & 0.562 ($n{=}48$) & 0.763--0.770 \\
where evidence merges (kept) & 0.807 ($n{=}140$) & 0.667 ($n{=}12$) & \textbf{0.796} \\
\bottomrule
\end{tabular}
\end{table}

\paragraph{UltraHorizon.}
The kept commit gate replays every logged observation of a rule class against the drafted rule and
withholds commitment while a replay fails, up to three times. It was added in the second admitted
bundle, alongside a per-class confound matrix and a probe plan (development 30.0 vs 16.7;
validation 33.3 vs 26.7). On a diagnostic run of 15 frozen episodes on unseen seeds of the
development environment, it held answers in 12 of the 15. It is off in the biology
environment, which has no machine-checkable rules to replay. The same
loop refused other self-verifications at the commit point. An earlier form of the commit gate was
tested in an earlier program with its own six development episodes. Used alone, it let the agent
commit in only 2 of 6, because the agent could not write rules the gate would accept. With
observation tables beside it, the agent committed in 5 of 6, and a clean re-measurement tied the
baseline at 36.0. A bundle with rival-slate and withheld-row replay lost 26.7 points on validation.
A sufficiency gate held answers in every development episode and lost 10 points. The development
record summarizes these results as ``verification rejects, it cannot identify'': a gate supports progress only
when it has evidence that separates rival rules.

\paragraph{Terminal-Bench.}
Earlier loops tried a separate auditor of finished work, a tool-using auditor executing adversarial
cases, an adversary writing failing probes against delivered code, a read-only non-author
reviewer, and checks written and graded by the agent itself. The variant without localization
(Table~\ref{tab:ablation}, fifth row) also admitted deterministic constraint checks run after each
call. Several of these did real work. The audits made two fixes that the grader confirmed. The
tool-using auditor raised one development task to 13 to 15 of 15 checks against at most 11 for
Claude Code, but stayed within noise across six tasks at 3.2 times the tokens. The adversary on
delivered code raised two development tasks, at 18 to 43 times the baseline's tokens. The
self-graded checks gave reproducible gains on two tasks while losing checks elsewhere. None was
admitted. The development record states why: ``the author cannot grade its own adversary: every
self-graded check mirrored the fix, fed the code's own formula as EXPECTED, or re-labelled a failing
check as a test artefact; a non-author reviewer found real bugs and moved no hidden check.'' These
runs led the loop to self-verification backed by an independent draw or a deterministic hook. A second
independent draw with a reconcile step ran out of budget in most draws, so that mechanism was not
tested fairly. Hooks that fired once at the agent's stop acted as advice. The admitted edit gate is
a hook that acts before the first edit of a shipped file. In development, the repair procedure
already had the agent write the invariant and probe first, so the gate had nothing to refuse. The gate
did not fire on two tasks whose edits went through scripts, and development attribution judged it
harmful on one.

\subsection{Self-verification switched off at test}
\label{app:verification-ablation}

Table~\ref{tab:verification-ablation} runs two frozen harnesses on their test sets with
self-verification switched off and everything else unchanged (Gemini~3.7 Flash, three seeds). On
UltraHorizon this removes the commit gate's rule replay; on LoCoMo, the claim judge and span checks.
The UltraHorizon run covers the 21 grid episodes of each test seed, and the LoCoMo run was judged
separately from Table~\ref{tab:memory}. Removing self-verification lowered the score at every seed in
both settings, by about half of Vestrum's gain over GAM.

\begin{table}[h]
\caption{Self-verification switched off at test (Gemini~3.7 Flash). UltraHorizon: mean score per
100 over the 21 grid episodes of each test seed. LoCoMo: strict accuracy over 956 probes.
$\pm$: standard deviation over seeds.}
\label{tab:verification-ablation}
\centering
\small
\begin{tabular}{llcccc}
\toprule
Setting & System & Seed 1 & Seed 2 & Seed 3 & Mean \\
\midrule
UltraHorizon (grid) & Vestrum & 69.52 & 68.57 & 74.29 & \textbf{70.79} $\pm$ 3.07 \\
 & without self-verification & 65.71 & 66.67 & 60.00 & 64.13 $\pm$ 3.61 \\
 & GAM & 58.10 & 56.19 & 55.24 & 56.51 $\pm$ 1.46 \\
\midrule
LoCoMo & Vestrum & 0.742 & 0.755 & 0.756 & \textbf{0.751} $\pm$ 0.008 \\
 & without self-verification & 0.733 & 0.737 & 0.727 & 0.732 $\pm$ 0.005 \\
 & GAM & 0.715 & 0.707 & 0.723 & 0.715 $\pm$ 0.008 \\
\bottomrule
\end{tabular}
\end{table}

\section{Figures}
\label{app:figures}

\begin{figure}[htbp]
\centering
\includegraphics[width=\linewidth]{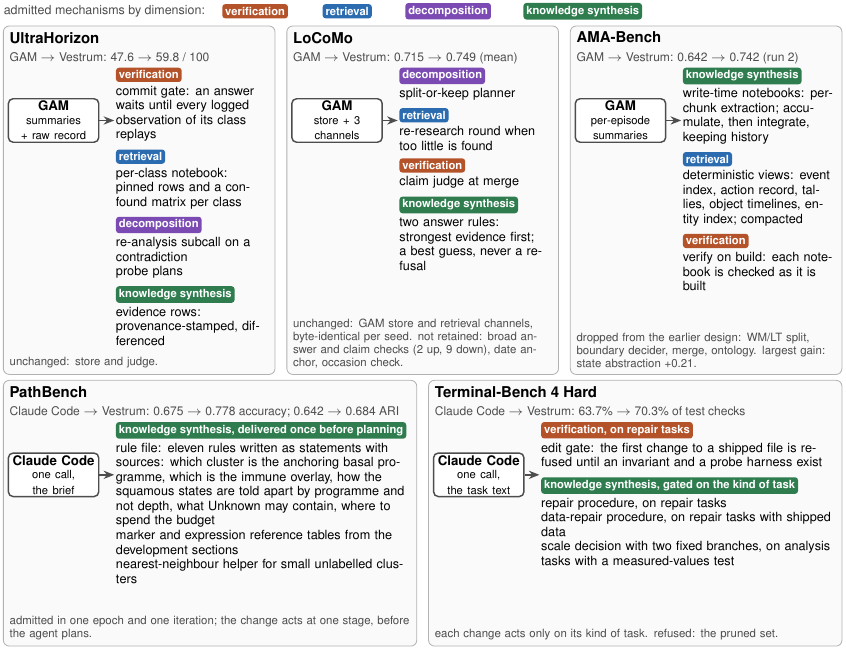}
\caption{The admitted harnesses. The top row starts from GAM and the bottom row from
Claude Code. Each panel shows the mechanisms by design dimension, with retained context and
earlier alternatives below. Scores are for whole harnesses.}
\label{fig:architectures}
\end{figure}

Figure~\ref{fig:architectures} shows the admitted harness in each setting, with mechanisms
colored by design dimension. Scores summarize the evaluated bundles. LoCoMo reports
three-replicate means. AMA-Bench reports the second answer-generation run, as in the upper block
of Table~\ref{tab:memory}.

\end{document}